\documentclass[hidelinks]{article}
\usepackage{spconf,amsmath,graphicx}
\usepackage[english]{babel}

\usepackage{subcaption}
\usepackage{hyperref}
\hypersetup{
    colorlinks=false,
    linkcolor=blue,
    filecolor=magenta,      
    urlcolor=blue,    
    }
\usepackage[final]{microtype}
\usepackage{xcolor}

\usepackage{enumitem}
\setlist{nosep, leftmargin=14pt}

\usepackage{placeins}

\renewcommand{\paragraph}[1]{\noindent\textbf{#1}\;}

\newcommand{\smallAddress}[1]{\vspace{-1.1mm}{\small #1}}

\def\sectionVspacePre{\vspace{-4mm}}
\def\sectionVspacePost{\vspace{-2mm}}
\def\sectionVspacePostExtra{\vspace{-2mm}}

\def\subsectionVspacePre{\vspace{-1mm}}
\def\subsectionVspacePreExtra{\vspace{-3mm}}  
\def\subsectionVspacePost{\vspace{-2mm}}

\def\mathCondensation{-1.2mm}

\title{Image Harmonization using Robust Restricted CDF Matching 
}
\name{%
\begin{tabular}{@{}c@{}}
    Roman Stoklasa$^{}$ \quad
    \thanks{This project was partially funded by 
    the Ministry of Health of the Czech Republic 
    (Grant No. NU21-08-00359) and the Ministry of Education, Youth and Sports of the Czech Republic (Project No. LM2023050).}
\vspace{-3mm}
\end{tabular}}

\address{%
  \smallAddress{$^{}$ Centre for Biomedical Image Analysis, Faculty of Informatics, Masaryk University, Brno, Czech Republic}
}

\begin{document}
%
\maketitle
%
%
\begin{abstract}


Deployment of machine learning algorithms into real-world practice is still a difficult task. 
One of the challenges lies in the unpredictable variability of input data, which may differ significantly among individual users, institutions, scanners, etc.
The input data variability can be decreased by using suitable data preprocessing with robust data harmonization. 
In this paper, we present a method of image harmonization using Cumulative Distribution Function (CDF) matching based on curve fitting. 
This approach does not ruin local variability and individual important features. 
The transformation of image intensities is non-linear but still  ``smooth and elastic", as compared to other known histogram matching algorithms. 
Non-linear transformation allows for a very good match to the template. 
At the same time, elasticity constraints help to preserve local variability among individual inputs, which may encode important features for subsequent machine-learning processing. 
The pre-defined template CDF offers a better and more intuitive control for the input data transformation compared to other methods, especially ML-based ones. 
Even though we demonstrate our method for MRI images, the method is generic enough to apply to other types of imaging data.

\end{abstract}
%
%
%
\begin{keywords}
harmonization, normalization, 
CDF, histogram matching, 
federated learning, FL, MRI, brain, tumor
\end{keywords}
%


\sectionVspacePre
\section{Introduction}%
\label{sec:intro}%
\sectionVspacePost\sectionVspacePostExtra

Medical imaging, particularly magnetic resonance imaging (MRI), 
has become an indispensable tool in clinical practice and research, 
providing detailed insights into anatomical structures and physiological processes. 
However, the variability in MRI data arising from differences in acquisition protocols, 
scanner hardware, or cross-institutional differences, pose significant challenges 
for accurate interpretation and robust automated processing of images.
This is especially significant in the Federated Learning environment, 
where a single model must adapt to a very different (and often 
non-iid: not independent and identically distributed) 
sets of images. 
Moreover, MRI images are created in arbitrary units of intensity, 
which are often incompatible between scanners, or even protocols.

Existing MRI harmonization techniques primarily fall into 
three categories: 
statistical approaches, intensity normalization, and machine learning methods. 
Statistical methods like ComBat \cite{reynolds2023combat}, 
originally proposed for gene expression microarrays \cite{johnson2007ComBat},
leverage empirical Bayes frameworks to adjust for batch 
effects in imaging data, effectively reducing variability across 
different scanners and acquisition protocols. 

Intensity normalization techniques, such as histogram matching and 
Z-score normalization, are widely used to align the intensity 
distributions of images from different sources.
Z-score normalization (also known as standardization), 
transforms the data so that the distribution of intensities 
has a mean of 0 and a standard deviation of 1.
Histogram matching \cite{nyul2000GeneralHistMatching} adjusts 
the intensity distribution of an image 
to resemble that of a reference dataset or reference distribution.
There have been multiple proposed algorithms, 
e.g., White Stripe harmonization \cite{shinohara2014WhiteStripe}, 
RAVEL \cite{fortin2016RAVEL}, mica \cite{Wrobel2020Mica} 
or RIDA \cite{RIDA2022}.

Machine learning approaches have gained traction in recent years, 
with techniques such as generative adversarial networks (GANs) 
offering innovative solutions for harmonizing images across varying 
conditions. While these methods can learn complex 
mappings between different imaging domains, their reliance on large 
datasets and extensive training can be limiting. 
Also, these methods are more like black boxes, which are hard to adjust, 
and there is no guarantee that some ``unexpected" input in the future 
would not produce incorrect result. 
Some of the recently published approaches include 
style-transfer techniques \cite{liu2023styleTransfer}, and methods 
like CALAMITI \cite{zuo2021CALAMITI}, 
MISPEL \cite{torbati2023mispel} and ESPA \cite{ESPA2024}.

In contrast to other recent methods that focus on modeling and/or correcting 
the \emph{scanner effect}, or which study the harmonization from more theoretical 
point of view, our aim is clearly different: we want to present 
a technically sound and robust method, still intuitive and straightforward, 
that would help data engineers in their practical situations -- to enhance 
the performance of their developed ML algorithms.
It should work not only in the experimental setups, like many other publications 
present, but mainly in real situations with real limitations and real problems with 
the data -- it should work out of the box for both healthy and pathological cases. 
We shall not forget about the big challenges when ML software is being deployed 
for testing in different clinical practices.
It is also very important to harmonize various channels -- including  
the structural (T1, T2, FLAIR, T1ce) and also diffusion channels 
(like IVIM and DTI) \cite{KOPRIVOVA2024-diffusion-review}. 
Since typical ML solutions combine various channels together, it is important
not only to harmonize these channels individually, but also to adjust 
their relative intensity ranges to avoid bias in the training. 

Other commonly used techniques often fail to achieve this. 
For example, the simple normalization by \textbf{percentile stretching} 
is not robust, since images may have significantly 
different ``weight" of its distribution tails. 
Thus, it is very difficult (if not impossible) to find a suitable percentile 
that works for all images without manual interventions 
-- leading to significant variations in the harmonized output. 
The \textbf{Z-score normalization} is also used very often
thanks to its simplicity, but it suffers from the fact
that a typical MRI image does not follow the normal distribution. 
The asymmetry of MRI distribution causes that most of the values 
are closely around 0, but there is still a significant amount of values 
in the "heavy tail". This may cause problems with the limited capability 
of neural networks to understand large dynamic ranges correctly. 
When the \textbf{histogram matching} is performed in a sub-optimal way \cite{fortin2016RAVEL,Wrobel2020Mica,RIDA2022}, 
this procedure may ruin the features in the image that may be important 
for further ML processing steps. In our proposal, as described below, we imply 
restrictions on how the CDF can be transformed to match the template, 
which allows the preservation of local variability among individual input images.
Despite the \textbf{Deep Learning and ML-based} methods being 'trendy' now, 
the harmonization methods based on this approach seem to be unnecessarily 
complex and complicated, prone to over-correction 
or alternation of the depicted structures themselves. 
They are often limited only to channels which they were trained for, 
not to mention their practical limitations when they come into 
the ``wild" of clinical practice, where unseen, unexpected, and non-iid data emerge.

In this paper, we introduce our version of analytical harmonization method 
based on constrained histogram matching. 
In our approach, the process of matching the image CDF to the template is 
solved as a curve-fitting optimization problem with specific constraints 
on how the image CDF can be transformed. 
By employing mathematical modeling techniques that are independent of deep machine learning, 
our method aims to enhance the robustness and reliability of subsequent image analyses 
and further processing.



\sectionVspacePre%
\section{Method}%
\label{sec:methods}%
\sectionVspacePost%

The main idea of our approach is to match the image CDF function to 
a pre-defined template CDF (target). 
The main difference from other histogram matching methods is that 
we solve the CDF-matching as a curve-fitting problem with post-processing
to shrink the long tails of the distribution.  
We are not modifying the histogram to ``perfectly" match the template CDF 
as in standard histogram matching approaches. 

Using the curve-fitting optimization brings multiple advantages. 
The image CDF and the template are ``loosely coupled" 
which allows for small local deviations in distribution to be preserved.
These deviations are typically signs of local features or abnormalities 
(e.g., pathologies) and shall not be suppressed because  
they may be important for further processing (e.g., brain tumor 
segmentation or classification). 
Moreover, the weak relation between image CDF and the template 
allows the template to be defined almost arbitrarily. 
The template CDF does not need to correspond precisely to the 
image distribution -- if the template is significantly 
different from image CDF, the only penalty we pay 
is that the curve fitting may lose some of its robustness; 
otherwise, this method would still work. 
The preparation of template CDF is independent of the 
process for intensity harmonization, as we will show below.

\begin{figure}[t]
    \centering
    \includegraphics[width=0.484\textwidth,height=60mm]{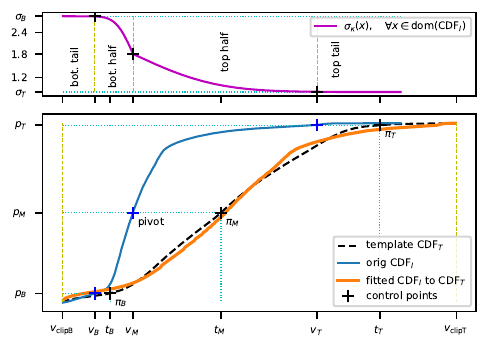}\vspace{-4mm}%
    \caption{ 
    Illustration of the main concepts for dual-scaling and 
    CDF-matching using curve fitting -- we fit $\text{CDF}_I$ to match the $\text{CDF}_T$.
    \vspace{-5mm}
    }
    \label{fig:basic-idea}
\end{figure}


During the curve fitting phase, we aim to modify the CDF function 
``elastically" -- we try to avoid any sudden changes, i.e., compression 
or stretching of local intensity ranges. 
Thus, we allow only two transformations of the image CDF: 
i) the smooth ``dual-scaling" defined with 
two scaling factors, and ii) uniform shift by a constant offset. 

In the following subsections, we describe in more detail all the 
individual steps: i) smooth dual-scaling, ii) tail shrinking, 
iii) how we derive template CDF, and iv) how the image intensity 
transformation is finally performed.

\subsectionVspacePre
\subsection{Smooth Dual-Scaling}
\subsectionVspacePost

The motivation for the ``smooth dual-scaling" transformation is that 
we want to scale CDF curve with two different scaling 
factors $\sigma_B$ and $\sigma_T$ along the x-axis: 
$\sigma_B$ applied to the bottom (left) half of the curve, 
and $\sigma_T$ to the top (right) part (see Fig.~\ref{fig:basic-idea}). 
Let's call \emph{pivot} the point that splits CDF curve 
into bottom and top parts. 
If the two scaling factors are applied directly, depending on 
their difference, this may create an unwanted sharp artifact 
at the pivot point.
Therefore, we define a point-wise scaling function $\sigma_\kappa(x)$ 
that smoothly blends between $\sigma_B$ and $\sigma_T$ using 
sigmoidal Error Function (erf). 
In the following equations, $x$ denotes the original intensity value. 
For given three intensity values $v_B < v_M < v_T$, we define: 
    \vspace{\mathCondensation}\[
    \Bar{x} = \text{interp1d}(x, \kappa, \epsilon)
    \vspace{\mathCondensation}\]
where $\kappa = (v_B, v_M, v_T)$ and $\epsilon = (-2, 0, 2)$ are 
x- and y-coordinates of control points 
defining piecewise-linear 1D interpolation of function $\text{interp1d}(\cdot)$. 
We are effectively mapping intensity values $x$ 
from interval $\langle v_B, v_T \rangle$ into $\text{erf}(\cdot)$ domain 
$\langle -2, 2 \rangle$, where its value ranges in interval
$\langle -0.995, 0.995 \rangle$. 
Then, the sigmoidal blending function $\beta_\kappa$ is defined as: 
    \vspace{\mathCondensation}\[
    \beta_\kappa(x) = 1 - \frac{1}{2}(\text{erf}(\Bar{x}) + 1) 
    \vspace{\mathCondensation}\]
which is used in the point-wise dual-scaling function 
$\sigma_\kappa(\cdot)$:
    \vspace{\mathCondensation}\[
    \sigma_\kappa(x, \sigma_B, \sigma_T) = \beta_\kappa(x) \cdot \sigma_B  + (1 - \beta_\kappa(x)) \cdot \sigma_T 
    \vspace{\mathCondensation}\]
The dual-scaling function is then used to transform 
original image intensities $x$ into modified ones using 
the $LUT_\text{ds}(x)$ function: 
\vspace{-3mm}\begin{equation}\label{eq:intensity-transform}
    LUT_\text{ds}(x) = (x - v_M) \cdot \sigma_\kappa(x, \sigma_B, \sigma_T) + \gamma
\end{equation}
where $\gamma$ is the uniform shift along the x-axis of CDF. 
Please note, that if $\sigma_B = \sigma_T$, this method 
naturally leads to the solution with uniform scaling.


\subsectionVspacePre\subsectionVspacePreExtra%
\subsection{Tail Shrinking}%
\subsectionVspacePost%

In many practical situations, we want to constrain possible 
intensity values to be within a given range, e.g., because 
we want to save the values as 8-bit or 12-bit integers. 
If the scaling and shift are applied to the original CDF, 
it may happen that its tails (either bottom or upper) 
may extend beyond the desired range. To counteract this problem, 
we introduce a ``tail shrinking" transformation. 
Here, we define the operation only on the upper (top) tail, 
because the shrinking of the bottom tail can be performed 
using the complementary function. 

\begin{figure}[t]
    \centering%
    \includegraphics[width=0.484\textwidth]{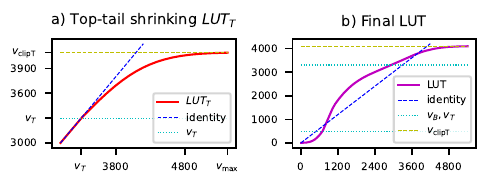}\vspace{-5mm}%
    \caption{ 
        Illustration of a) top-tail shrinking $LUT_T$ function, and 
        b) an example of final intensity mapping from the original 
        intensity range $\langle 5, 5418 \rangle$ 
        to the normalized 12-bit range $\langle 1, 4095 \rangle$. 
    \vspace{-7mm}
    }
    \label{fig:LUT-examples}
\end{figure}

Let $v_{\max}$ be the highest intensity in the image, 
$v_\text{clipT} < v_{\max}$ be the top clipping value, 
and
$v_T < v_\text{clipT}$ be the defined top intensity value, which designates 
where the upper tail of CDF starts. 
Then, we can define the upper-tail intensity transformation function 
$LUT_T(\cdot)$ as:
\vspace{\mathCondensation}\[
LUT_T(x, v_T, v_{\max}, v_{\text{clipT}}) = v_T + r_T \cdot \text{erf}\left( 2\frac{x-v_T}{r_S} \right)
\vspace{\mathCondensation}\]
where $r_S$ and $r_T$ are ranges for source and target tail 
intensities respectively: 
\vspace{\mathCondensation}\begin{align*}
r_S &= v_{\max} - v_T   &
r_T &= v_\text{clip} - v_T
\vspace{\mathCondensation}\end{align*}
The top-tail shrinking $LUT_T$ function is visually demonstrated in 
Fig.~\ref{fig:LUT-examples}a.
To shrink the tail on the bottom side, we can utilize the following formulas: 
    \vspace{\mathCondensation}\begin{align*}
    x' &= v_{\max} - x   &   v_B' &= v_{\max} - v_B \\ 
    v_{\min}' &= v_{\max} - v_{\min}     &   v_\text{clipB}' &= v_{\max} - v_\text{clipB} 
    \vspace{\mathCondensation}\end{align*}%
\vspace{-5mm}\[
    LUT_B(x) = v_{\max} - LUT_T(x', v_B', v_{\min}', v_\text{clipB}')
\]

\subsectionVspacePre\subsectionVspacePreExtra%
\subsection{Template CDF preparation}%
\subsectionVspacePost%

\begin{figure*}[t]
    \centering
    \includegraphics[width=0.99\textwidth,height=40mm]{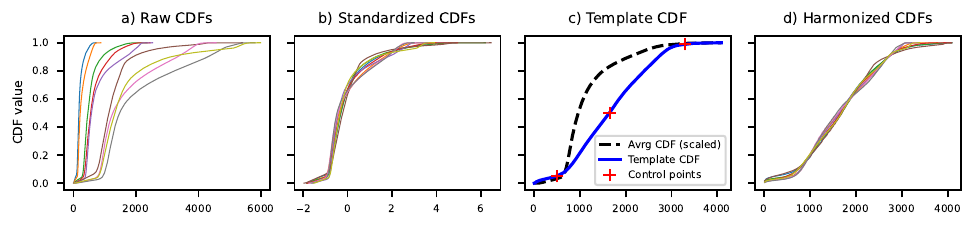}\vspace{-4mm}%
    \caption{ 
    An example of the harmonization process for 9 MRI T2 images. 
    a) CDF functions of raw images, 
    b) CDF functions after z-score normalization, 
    c) template $\text{CDF}_T$ fitted to three control points  
    (configured to utilize full 12-bit range) and comparison with the average CDF,
    d) CDF functions of harmonized images (please
    note the desired tiny local deviations from the $\text{CDF}_T$).
    \vspace{-5mm}
    }
    \label{fig:tgt-cdf}
\end{figure*}

As we mentioned before, the template CDF curve 
can be defined independently from this harmonization method. 
Here, we demonstrate our solution for how we derive 
average CDF computed from 
a set of examples (typically, some training examples we have available).
We just need to pay attention to the fact that these training 
images are typically in the raw form before any harmonization is applied. 

To derive average CDF from potentially highly variable samples, 
we use a little trick: we apply Z-score normalization  
to the images first and derive CDF curves from such standardized distributions. 
At this stage, the potential variability of Z-score normalization
is not a problem, because we will average many CDFs together, so 
slight variations will cancel out. 
We obtain one average CDF curve: $\text{CDF}_A$.

Then, we define three control points: bottom $\pi_B = (t_B, p_B)$, 
middle $\pi_M = (t_M, p_M)$, and top $\pi_T = (t_T, p_T)$.
Each point is defined with the target percentile ($p_B, p_M, p_T$) 
and the target intensity value ($t_B, t_M, t_T$).
These three points then serve as the key points to which we fit 
the $\text{CDF}_A$ curve. Thanks to the dual-scaling, we can always find 
a suitable fit (see Fig.~\ref{fig:tgt-cdf}). 
The last step is to apply tail-shrinking to fit all 
intensity values into the desired range if there is such a requirement.  

Typically, we create a separate template CDF for every channel, 
because the distributions of individual channels differ significantly. 
However, the decisive key points $\pi_B$, $\pi_M$, and $\pi_T$ can be 
reused as they rarely require precise tuning.

\subsectionVspacePre\subsectionVspacePreExtra%
\subsection{Image transformation}%
\subsectionVspacePost%

\begin{figure}[tb]
    \centering%
    \includegraphics[width=0.484\textwidth]{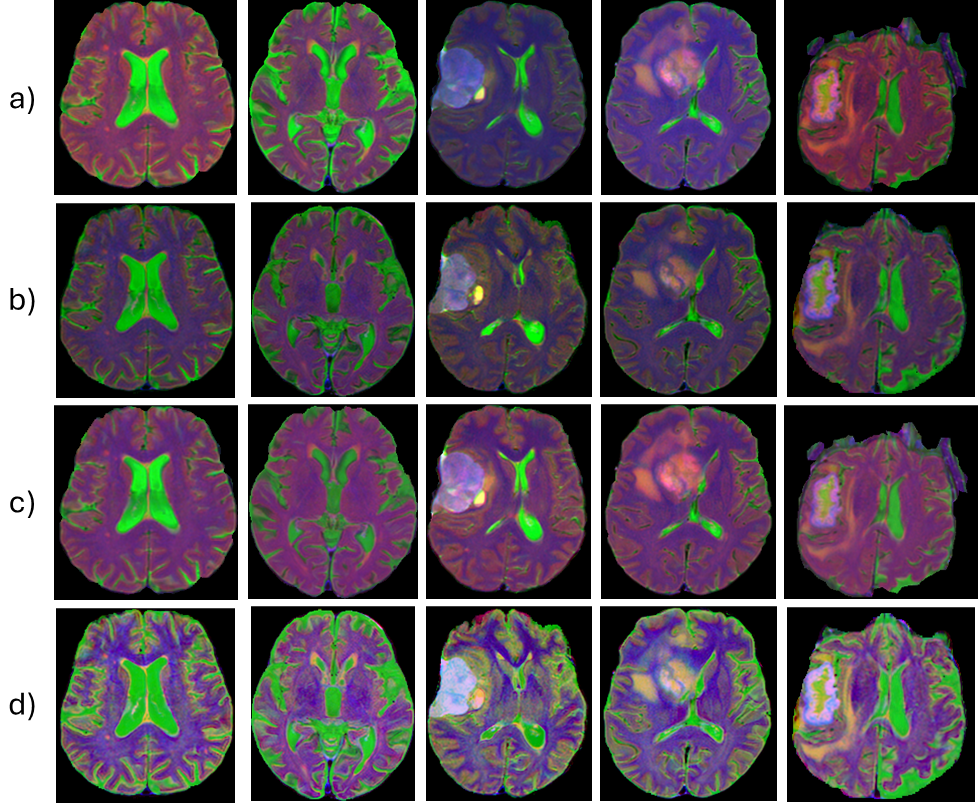}\vspace{-2mm}%
    \caption{%
    An example of harmonized results of 5 patients (cols) and different harmonization algorithms: 
    a) percentile stretch, b) z-score normalization, 
    c) our CDF-matching with 2 control points $\pi_B, \pi_T$, 
    d) our CDF-matching with all 3 control points yielding enhanced contrast. 
    MRI channels FLAIR, T2 and T1ce are combined as RGB color. 
    \vspace{-5mm}
    }%
    \label{fig:mri-brain-comparison}%
\end{figure}

Once we explained all the necessary tools above, 
the description of actual image harmonization is very easy. 
Let's assume we have already defined template $\text{CDF}_T$ 
and the control points $\pi_B, \pi_M, \pi_T$. 
For any input image $I$, we can perform the transformation of 
its intensities with the following recipe:
\begin{enumerate}
    \item Compute the image $\text{CDF}_I$ for input image $I$.
    \item Fit $\text{CDF}_I$ to $\text{CDF}_T$ using curve fitting optimization, where we allow transformation of $\text{CDF}_I$ in terms of double-scaling and uniform shifting. A fitted solution will give us optimal values of parameters $\sigma_B, \sigma_T$, and $\gamma$.
    \item For obtained parameters $\sigma_B, \sigma_T$, and $\gamma$, 
    use $LUT_\text{ds}$ function from Eq.~\ref{eq:intensity-transform} 
    to transform the image intensities. 
    \item Apply tail-shrinking $LUT_T$ and $LUT_B$ if the concrete application requires constrained intensity values.  
\end{enumerate}
Our implementation for Python is available in 
GitLab\footnote{\url{https://gitlab.fi.muni.cz/cbia/libs-public/cdf-normalizer}}.
Please note that this process actually solves two tasks simultaneously:
i) it harmonizes images because their intensity ranges will closely follow
the template $\text{CDF}_T$, and ii) it can increase contrast in images 
if the template $\text{CDF}_T$ is configured suitably such that 
it utilizes most of the available intensity range.


\sectionVspacePre%
\section{Experiments and Results}%
\label{sec:results}
\sectionVspacePost%



We tested this method extensively for MRI images of the brain, 
both for patients with tumors (256) and healthy patients (56). 
When we employed this harmonization algorithm for our 
Brain MRI Screening Tool \cite{stoklasa2024ScreeningTool}, 
which works in federated environment, the precision of the 
tool significantly increased (see Table~\ref{tab:results}). 
The model was then tested on 102 examinations. We are
reporting on four main metrics: 
i) the Exam-averaged Dice coefficient (EaD) -- the average 
of Dice coeffs. computed for each MRI exam separately; 
ii) the Global Dice (G.Dice) coefficient -- computed
over the combined volume of all examinations together 
(i.e., as if all images would be concatenated into one huge 3D image); 
iii) sensitivity (recall), 
and iv) precision.
In Fig.~\ref{fig:mri-brain-comparison}, we present a preview of 
5 patients harmonized with different methods as RGB images 
to demonstrate also relative relation between individual channels.
You can see that percentile stretch (a) is the least stable. 
Z-score normalization (b) has some subtle issues, especially 
with contrast and high values (see 3\textsuperscript{rd} and 4\textsuperscript{th} image). 
This trend can also be seen in Fig.~\ref{fig:tgt-cdf}b, 
where most of the values spread over a narrow area around 0, 
while the upper tail is still pretty heavy and extends to high values. 
Our method (depicted in rows \emph{c} and \emph{d}) produces 
the most homogeneous images, 
where the balance between individual channels is well-controlled 
and stable. 
Row \emph{d} demonstrate also the additional benefit of increased contrast
when the template CDF utilizes a larger portion of the intensity range. 
This result was achieved with parameters: $\pi_B = (0.1, 500), \pi_M = (0.5, 1650), \pi_T = (0.99, 3300), v_\text{clipB} = 1, v_\text{clipT} = 4095$ 
(i.e., normalizing to 12-bit).
Since we use value 0 as a special value for background in MRI images, 
we set $v_\text{clipB} = 1$ to avoid any interference. 

Setting the parameters $\pi_B, \pi_M, \pi_T$ depends on
a typical distribution you work with. 
Typically, set $\pi_M$ to define the intensity you want 
to be assigned to the median. 
Adjust points $\pi_B$ and $\pi_T$ such that you cover most of 
the target range (to increase the contrast), but still leave enough space 
where the tail values could be squeezed into via tail-shrinking 
(if you use the clipping values). 
These parameters primarily affect how template $\text{CDF}_T$ 
is created, and the method itself is not very sensitive 
to these values so they can be frequently reused.

\begin{table}[t]
    \centering
\caption{Performance of the FL-based Brain MRI Screening Tool \cite{stoklasa2024ScreeningTool} with our CDF-matching harmonization method, as compared to the results obtained by exactly the same network trained and tested on input images harmonized using simple percentile stretch.
\vspace{-2mm}
}
\label{tab:results}
    \begin{tabular}{|c||c|c|c|c|} \hline 
 Method used:& EaD & G-Dice & Sens. & Prec.  \\ \hline\hline
CDF-matching&  \textbf{0.837}& \textbf{0.884} & \textbf{0.98} & \textbf{0.91} \\ \hline 
Percentile stretch& 0.803&0.852& 0.94& 0.90\\\hline
    \end{tabular}
    \vspace{-5mm}
\end{table}

\sectionVspacePre%
\section{Conclusion}%
\label{sec:conclusion}%
\sectionVspacePost\sectionVspacePostExtra%

We presented our proposed image harmonization method based on the 
CDF matching using curve fitting. 
Curve fitting allows for local image deviations from template CDF, 
which i) helps preserve important image features for subsequent processing,
and ii) allows for lowering requirements imposed on the template CDF. 
We used this method extensively for MRI images with many different 
channels (structural and also diffusion-weighted), 
and it always worked exceptionally well. 
This method can be easily used in any project and the 
Python implementation is \href{https://gitlab.fi.muni.cz/cbia/libs-public/cdf-normalizer}{publicly available} (see footnote).


\FloatBarrier

\section{Acknowledgments}
\label{sec:acknowledgments}

The author would like to thank our colleagues 
M. Dostál, T. Kopřivová, and M. Keřkovský for 
providing MRI data on which this method was demonstrated,
and Prof. Michal Kozubek for his support of this development.

\bibliographystyle{IEEEbib}
\bibliography{references}

\begin{thebibliography}{10}

\bibitem{reynolds2023combat}
Maxwell Reynolds et~al.,
\newblock ``Combat harmonization: Empirical bayes versus fully bayes approaches,''
\newblock {\em NeuroImage: Clinical}, vol. 39, pp. 103472, 2023.

\bibitem{johnson2007ComBat}
W~Evan Johnson et~al.,
\newblock ``Adjusting batch effects in microarray expression data using empirical bayes methods,''
\newblock {\em Biostatistics}, vol. 8, no. 1, pp. 118--127, 2007.

\bibitem{nyul2000GeneralHistMatching}
L{\'a}szl{\'o}~G Ny{\'u}l et~al.,
\newblock ``New variants of a method of mri scale standardization,''
\newblock {\em IEEE transactions on medical imaging}, vol. 19, no. 2, pp. 143--150, 2000.

\bibitem{shinohara2014WhiteStripe}
Russell~T Shinohara et~al.,
\newblock ``Statistical normalization techniques for magnetic resonance imaging,''
\newblock {\em NeuroImage: Clinical}, vol. 6, pp. 9--19, 2014.

\bibitem{fortin2016RAVEL}
Jean-Philippe Fortin et~al.,
\newblock ``Removing inter-subject technical variability in magnetic resonance imaging studies,''
\newblock {\em NeuroImage}, vol. 132, pp. 198--212, 2016.

\bibitem{Wrobel2020Mica}
J.~Wrobel et~al.,
\newblock ``Intensity warping for multisite mri harmonization,''
\newblock {\em NeuroImage}, vol. 223, 2020.

\bibitem{RIDA2022}
Donatas Sederevičius et~al.,
\newblock ``A robust intensity distribution alignment for harmonization of t1w intensity values,''
\newblock {\em bioRxiv}, p. 2022.06.15.496227, 6 2022.

\bibitem{liu2023styleTransfer}
Mengting Liu et~al.,
\newblock ``Style transfer generative adversarial networks to harmonize multisite mri to a single reference image to avoid overcorrection,''
\newblock {\em Human Brain Mapping}, vol. 44, no. 14, pp. 4875--4892, 2023.

\bibitem{zuo2021CALAMITI}
Lianrui Zuo et~al.,
\newblock ``Unsupervised mr harmonization by learning disentangled representations using information bottleneck theory,''
\newblock {\em NeuroImage}, vol. 243, pp. 118569, 2021.

\bibitem{torbati2023mispel}
Mahbaneh~Eshaghzadeh Torbati et~al.,
\newblock ``Mispel: A supervised deep learning harmonization method for multi-scanner neuroimaging data,''
\newblock {\em Medical image analysis}, vol. 89, pp. 102926, 2023.

\bibitem{ESPA2024}
Mahbaneh Eshaghzadeh~Torbati et~al.,
\newblock ``Espa: An unsupervised harmonization framework viaÂ enhanced structure preserving augmentation,''
\newblock in {\em Medical Image Computing and Computer Assisted Intervention -- MICCAI 2024}, Cham, 2024, pp. 184--194, Springer Nature Switzerland.

\bibitem{KOPRIVOVA2024-diffusion-review}
Tereza Kopřivová, Miloš Keřkovský, Tomáš Jůza, Václav Vybíhal, Tomáš Rohan, Michal Kozubek, and Marek Dostál,
\newblock ``Possibilities of using multi-b-value diffusion magnetic resonance imaging for classification of brain lesions,''
\newblock {\em Academic Radiology}, vol. 31, no. 1, pp. 261--272, 2024.

\bibitem{stoklasa2024ScreeningTool}
Roman Stoklasa et~al.,
\newblock ``Brain mri screening tool with federated learning,''
\newblock in {\em 2024 IEEE International Symposium on Biomedical Imaging (ISBI)}. IEEE, 2024, pp. 1--5.

\end{thebibliography}

\sectionVspacePre%
\section{Compliance with ethical standards}%
\label{sec:ethics}%
\sectionVspacePost%

This is a mathematical study for which no ethical approval was required.
The MRI data used for the demonstration of this method come from 
the study, which was performed in line with the principles of the 
Declaration of Helsinki. 
Approval was granted by the Ethics Committee of University Hospital Brno.

\end{document}